\documentclass[]{sigplanconf}
\usepackage[utf8]{inputenc}
\usepackage{lmodern}
\usepackage{alltt} 
\usepackage{amsthm}
\usepackage{booktabs}
\usepackage{url}

\usepackage{xspace}
\usepackage{amsmath}

\def\holfour{\textsf{HOL4}\xspace}
\def\isabellehol{\textsf{Isabelle/HOL}\xspace}
\def\hollight{\textsf{HOL Light}\xspace}
\def\mizar{\textsf{Mizar}\xspace}

\def\vampire{\textsf{Vampire}\xspace}
\def\eprover{\textsf{E-prover}\xspace}
\def\zthree{\textsf{z3}\xspace}

\def\sml{\textsf{SML}\xspace}
\def\holyhammer{\textsf{HOL(y)Hammer}\xspace}
\def\sledgehammer{\textsf{Sledgehammer}\xspace}
\def\mizAR{\textsf{MizAR}\xspace}
\def\mizar{\textsf{Mizar}\xspace}

\def\metis{\textsf{Metis}\xspace}

\theoremstyle{remark}
\newtheorem{example}{Example} 
 
\begin{document}
\permissiontopublish
\conferenceinfo{CPP~'15}{January 13--14, 2015, Mumbai, India}
\copyrightyear{2015}
\copyrightdata{978-1-4503-3300-9/15/01}
\doi{2676724.2693173} 
\title{Premise Selection and External Provers for HOL4}

\authorinfo{Thibault Gauthier \and Cezary Kaliszyk}
           {University of Innsbruck}
           {\{thibault.gauthier,cezary.kaliszyk\}@uibk.ac.at}

\maketitle

\begin{abstract}
  Learning-assisted automated reasoning has recently gained popularity
  among the users of \isabellehol, \hollight, and \mizar. In this paper, we
  present an add-on to the \holfour proof assistant and an adaptation of the \holyhammer
  system that provides machine learning-based pre\-mise selection and automated
  reasoning also for \holfour. We efficiently record the \holfour dependencies and
  extract features from the theorem statements, which form a basis for
  premise selection. \holyhammer transforms the \holfour statements in the
  various TPTP-ATP proof formats, which are then processed by the ATPs.

  We discuss the different evaluation settings: ATPs, accessible lemmas,
  and premise numbers. We measure the performance of \holyhammer on the
  \holfour standard library. The results are combined accordingly and
  compared with the \hollight experiments, showing a comparably high
  quality of predictions. The system directly benefits \holfour users by
  automatically finding proofs dependencies that can be reconstructed
  by \metis.
\end{abstract}



\section{Introduction}
The \holfour proof assistant~\cite{hol4} provides its users with a full ML programming 
environment
in the LCF tradition. Its simple logical kernel and interactive interface allow safe and 
fast developments,
while the built-in decision procedures 
can automatically establish many simple theorems, leaving only the harder goals to its 
users.
However, manually proving theorems based on its simple rules is a tedious task. Therefore,
general purpose automation has been developed internally, based on model elimination
(\textsf{MESON}~\cite{meson}), tableau (\textsf{blast}~\cite{blast}), or resolution 
(\metis~\cite{metis}).
Although essential to \holfour developers, the methods are so far not able to compete with 
the external ATPs~\cite{eprover,KovacsCAV13} optimized for fast proof search with many 
axioms present and
continuously evaluated on the TPTP library~\cite{TPTP} and updated with the most successful 
techniques. The TPTP (Thousands of Problems for Theorem Provers) is a library of test problems for automated theorem proving (ATP) systems. This standard enables
convenient communication between different systems and researchers.

On the other hand, the \holfour system provides a functionality to search
the database for theorems that match a user chosen pattern. 
The search is semi-automatic and
the resulting lemmas are not necessarily helpful in proving the conjecture. An approach
that combines the two: searching for relevant theorems and using automated reasoning methods
to (pseudo-)minimize the set of premises necessary to solve the goal, forms the basis of
``hammer'' systems such as \sledgehammer~\cite{sledgehammer10} for \isabellehol,
\holyhammer~\cite{holyhammer} for \hollight or \mizAR for 
\mizar~\cite{DBLP:journals/corr/KaliszykU13b}. Furthermore, apart from syntactic 
similarity of a goal to known
facts, the relevance of a fact can be learned by analyzing dependencies in previous proofs
using machine learning techniques~\cite{malarea}, which leads to a significant increase
in the power of such systems~\cite{mash}.

In this paper, we adapt the \holyhammer system to the \holfour system and test its 
performance
on the \holfour standard library. 
The libraries of \holfour and \hollight are exported together with proof dependencies and
theorem statement features; the predictors learn from the dependencies and the features to
be able to produce lemmas relevant to a conjecture. Each problem is translated to the TPTP
FOF format. When an ATP finds a proof, the necessary premises are extracted. They are read
back to \holfour as proof advice and given to \metis for reconstruction.

An adapted version of the resulting software is made available to the users of \holfour in
interactive session, which can be used in newly developed theories. Given a conjecture, the
\sml function computes every step of the interaction loop and, if successful, returns the
conjecture as a theorem:

  
  \begin{example}(\holyhammer interactive call)  
  \begin{alltt}
  load "holyHammer"; 
    val it = (): unit  
  holyhammer ``1+1=2``;
    Relevant theorems: ALT\_ZERO ONE TWO ADD1
    metis: r[+0+6]\#
    val it = |- 1 + 1 = 2: thm
  \end{alltt}
  \end{example}

The \holfour prover already benefits from export to SMT solvers such as 
\textsf{Yices}~\cite{WeberTjark}, \textsf{Z3}~\cite{Z3reconstruction} and 
\textsf{Beagle}~\cite{tgckckmn-paar14-accepted}. These methods  perform best when solving 
problems from the supported theories of the SMT solver. Comparatively, \holyhammer is a 
general purpose tool as it relies on ATPs without theory reasoning and it can provide 
easily\footnote{reconstruction rate is typically above 90\%} re-provable problem to \metis.

The \holfour standard distribution has since long been equipped with proof recording
kernels \cite{Wong95recordingand,DBLP:conf/itp/KumarH12}. We first considered adapting 
these kernels for our aim. But as 
machine learning only needs the proof dependencies and the approach based on full proof 
recording is not efficient, we perform minimal modifications to the original kernel.

\paragraph{Contributions}
We provide learning assisted automated reasoning for \holfour and evaluate its performance in comparison to that in \hollight. In order to do so, we :

\begin{itemize}
\item Export the \holfour data
  
  Theorems, dependencies, and features are exported by
  a patched version of the \holfour kernel. It can record dependencies between theorems and keep
  track on how their conjunctions are handled along the proof. We export the \holfour standard
  libraries (58 types, 2305 constants, 11972 theorems) with respect to a strict name-space rule so
  that each object is uniquely identifiable, preserving if possible its original name.

\item Reprove

  We test the ability of a selection of external provers to reprove theorems from
  their dependencies.
\item Define accessibility relations

  We define and simulate different development environments, with
  different sets of accessible facts to prove a theorem.

\item Experiment with predictors

  Given a theorem and a accessibility relation, we use machine learning techniques to find
  relevant lemmas from the accessible sets. Next, we measure the quality of the predictions
  by running ATPs on the translated problems.

\end{itemize}


The rest of this paper is organized as follows. In Section~\ref{sec:palibs} we describe the export of the \holfour and \hollight data into a common format and the recording of dependencies in \holfour. In Section~\ref{sec:setting}, we present the different parameters: ATPs, proving environments, accessible sets, features, and predictions. We select some of them for our experiments and justify our choice.
In Section~\ref{sec:experiments} we present the results of the \holfour experiments, relate them to previous \holyhammer experiments and explain how this affects the users.
Finally in Section~\ref{sec:concl} we conclude and present an outlook on the future work.

\section{Sharing HOL data between \holfour, \hollight and \holyhammer}\label{sec:palibs}

In order to process \hollight and \holfour data in a uniform way in \holyhammer, we export objects
from their respective theories, as well as dependencies between theorems into a common format.
The export is available for any \holfour and \hollight development. We shortly describe the
common format used for exporting both libraries and present in more detail our methods for
efficiently recording objects (types, constants and theorems) and precise dependencies in \holfour.
We will refer to \holyhammer~\cite{holyhammer} for the details on recording objects and dependencies
for \hollight formalizations.

\hollight and \holfour share a common logic (higher-order logic with implicit shallow polymorphism),
however their implementations differ both in terms of the programming language used (\textsf{OCaml}
and \sml respectively), data structures used to represent the terms and theorems (higher-order
abstract syntax and de Bruijn indices respectively), and the exact inference rules provided by the
kernel. As \holyhammer has been initially implemented in \textsf{OCaml} as an extension of \hollight,
we need to export all the \holfour data and read it back into \holyhammer, replacing its type and
constant tables. The format that we chose is based on the TPTP THF0 format~\cite{tptpthf0} used by
higher-order ATPs. Since formulas contains polymorphic constants which is not supported by the THF0 format, we will present an experimental extension of this format where the type arguments of polymorphic constants are given explicitly.

\begin{example} (experimental template)
  \begin{alltt}
    tt(name, role, formula)  
  \end{alltt}
  The field name is the object's name. The field role is "ty" if the object is a constant 
  or a type, and "ax" if the object is a theorem. The field formula is an experimental 
  THF0 formula. 
\end{example}

\begin{example} (Object export from \holfour to an experimental format)
  \begin{itemize}
  \item Type
  \begin{alltt}
 (list,1) \(\rightarrow\) tt(list, ty, \$t > \$t).
\end{alltt}
\item Constant
  \begin{alltt}
(HD,``:'a list -> :'a``) \(\rightarrow\) 
  tt(HD, ty, ![A:\$t]: (list A > A).
(CONS,``:'a -> :'a list -> :'a list``) \(\rightarrow\) 
  tt(CONS ,ty, ![A:\$t]: (A > list A > list A).
  \end{alltt}
  \item Theorem
    \begin{alltt}
(HD,``\(\forall\) n:int\ t:list[int].\ HD (CONS n t) = n``) \(\rightarrow\) 
  tt(HD0, ax, (![n:int, t:(list int)]:
    ((HD int) ((CONS int) n t) = n).
    \end{alltt}
  \end{itemize}
In this example, \texttt{\$t} is the type of all basic types.
\end{example}
All names of objects are prefixed by a namespace identifier, that allow identifying the
prover and theory they have been defined in. For readability, the namespace prefixes have
been omitted in all examples in this paper.

\subsection{Creation of a \holfour theory}
In \holfour, types and constants can be created and deleted during
the development of a theory. These objects are named at the moment they are created.
A theorem is a \sml value of type $thm$ and can be derived from a set of basic rules, which is an instance of a typed higher-order classical logic.
To distinguish between important lemmas and theorems created by each small steps, the user can name and delete theorems (erase the name).
Each named object still present at the end of the development is saved and thus can be called in future theories.

   There are two ways in which an object can be lost in a theory: either it is deleted or 
   overwritten. As proof dependencies for machine learning get more accurate when more
   intermediate steps are available, we decided 
   to record all created objects, which results in the creation of slightly bigger 
   theories. As the originally saved objects can be called from other theories, their names 
   are preserved by our transformation. Each lost object whose
   given name conflicts with the name of a saved object of the same type is renamed.

\paragraph{Deleted objects}
 The possibility of deleting an object or even a theory is mainly here to hide internal 
 steps or to make the theory look nicer. We chose to remove this possibility by canceling 
 the effects of the deleting functions. This is the only user-visible feature that behaves
  differently  in our dependency recording kernel.

\paragraph{Overwritten objects}  
  An object may be overwritten in the development. As we prevent objects from being deleted,
  the likelihood of this happening is increased. This typically happens when a generalized version of a theorem is proved and is given the same name as the initial theorem.
   In the case of types and constants, the internal \holfour mechanism already 
  renames overwritten objects. Conversely, theorems are really erased. To avoid
  dependencies to theorems that have been overwritten, we automatically rename the theorems
  that are about to be overwritten.

\subsection{Recording dependencies}
  Dependencies are an essential part of machine learning for theorem proving, as 
  they provide the examples on which predictors can be trained. We focus on 
  recording dependencies between named theorems, since they are directly accessible to a 
  user. The time mark of our method slows down the application of any rules by a 
  negligible amount.

Since the statements of 951 \holfour theorems are conjunctions, sometimes
consisting of many toplevel conjuncts, we have refined our method to record dependencies
between the toplevel conjuncts of named theorems.

\begin{example}(Dependencies between conjunctions)
\begin{alltt}
ADD\_CLAUSES: 0 + m = m \(\wedge\) m + 0 = m \(\wedge\) 
SUC m + n = SUC (m + n) \(\wedge\) m + SUC n = SUC (m + n)

ADD_ASSOC depends on:
  ADD_CLAUSES_c1: 0 + m = m
  ADD_CLAUSES_c3: SUC m + n = SUC (m + n)
  ...
\end{alltt}
The conjunct identifiers of a named theorem \texttt{T} are noted \texttt{T\_c1}, $\ldots$, \texttt{T\_cN}. 
\end{example}
In certain theorems, a toplevel universal quantifier shares
a number of conjuncts. We will also split the conjunctions
in such cases recursively. This type of theorem is less
frequent in the standard library (203 theorems).

\begin{example} (Conjunctions under quantifier)
\begin{alltt}
MIN_0: \(\forall\) n. (MIN n 0 = 0) \(\wedge\) (MIN 0 n = 0)
\end{alltt}
\end{example}

By splitting conjunctions we expect to make
the dependencies used as training examples for machine learning
more precise in two directions. First, even if a theorem is too
hard to prove for the ATPs, some of its conjuncts might be provable.
Second, if a theorem depends on a big conjunction, it typically depends
only on some of its conjuncts. Even if the precise conjuncts are
not clear from the human-proof, the reproving methods can often
minimize the used conjuncts. Furthermore, reducing the number of
conjuncts should ease the reconstruction.

\subsection{Implementation of the recording}\label{sec:record}
The \holfour type of theorems $thm$ includes a tag field in order to remember which oracles
and axioms were necessary to prove a theorem. Each call to an oracle or axiom creates a
theorem with the 
   associated tag. When applying a rule, all oracles and axioms from the tag of the parents 
   are respectively merged, and given to the conclusion of the rule.
   To record the dependencies, we added a third field to the tag,
   which consists of a dependency identifier and its dependencies.

 \begin{example}\label{exampletag} (Modified tag type)
  \begin{alltt}  
  type tag = ((dependency_id, dependencies), 
              oracles, axioms)
  type thm = (tag, hypotheses, conclusion)
  \end{alltt}
 \end{example}

   Since the name of a theorem may change when it is overwritten, we create unmodifiable 
   unique identifiers at the moment a theorem is named. 
   
   It consists of the name of the  
   current theory and the number of previously named theorems in this theory. As a
   side effect, this enables us to know the order in which theorems are named which is compatible by construction with the pre-order given by the dependencies. 
   Every variable of type $thm$ which is not named is given the identifier $unnamed$. Only identifiers of named theorems will appear in the dependencies.

   We have implemented two versions of the dependency recording  algorithm, one that tracks
   the dependencies between
   named theorems, other one tracking dependencies between their conjuncts.
   For the named theorems, the dependencies are a set of identified theorems used to prove the 
   theorem. The recording is done by specifying how each rule creates the tag of the 
   conclusion from the tag of its premises. The dependencies of the conclusion 
   are the union of the dependencies of the unnamed premises with its named premises.
    
   This is achieved by a simple modification of the \texttt{Tag.merge} function already applied to the tags of the premises in each rule.


When a theorem $\vdash A\,\wedge\,B$ is derived from the theorems $\vdash A$ and $\vdash B$,
the previously described algorithm would make the dependencies of this theorem the union
of the dependencies of the two. If later other theorems refer to it, they will get the union
as their dependencies, even if only one conjunct contributes to the proof. In this subsection
we define some heuristics that allow more precise tracking of dependencies of the conjuncts
of the theorems.

In order to record the dependencies between the conjuncts, we do not record the conjuncts of named theorems,
but only store their dependencies in the tags.  The dependencies are represented as a tree, in which each
leaf is a set of conjunct identifiers (identifier and the conjunct's address). Each leaf of the tree represents the respective conjunct $c_i$ in the
theorem tree and each conjunct identifier represents a conjunct of a named
goal to prove $c_i$.

\begin{example} (An example of a theorem and its dependencies)
\begin{alltt}
Th0 (named theorem): A \(\wedge\) B
Th1: C \(\wedge\) (D \(\wedge\) E) 
     with dependency tree Tree([Th0],[Th0_c2])

This encodes the fact that:
  C depends on Th0. 
  D \(\wedge\) E depends Th0_c2 which is B.
  

\end{alltt}
\end{example}

Dependencies are combined at each inference rule application and dependencies will contain 
only conjunct identifiers. If not specified, a premise will pass on its identifier if it is a named conjunct 
(conjunct of a named theorem) and its dependency tree otherwise. We call such trees passed 
dependencies. The idea is that the dependencies of a named conjunct should not transmit 
its dependencies to its children but itself. Indeed, we want to record the direct 
dependencies and not the transitive ones.

For rules that do not preserve the structure of conjunctions, we flatten the dependencies, i.e. we return a root tree containing the set of all (conjunct) identifiers in the passed dependencies.
We additionally treat specially the rules used for the top level organization of conjunctions: \texttt{CONJ},
\texttt{CONJUNCT1}, \texttt{CONJUNCT2}, \texttt{GEN}, \texttt{SPEC}, and \texttt{SUBST}.

 \begin{itemize}  
  \item \texttt{CONJ}:
    It returns a tree with two branches, consisting of the passed dependencies of its 
    first and second premise.
 \item \texttt{CONJUNCT1} (\texttt{CONJUNCT2}):
   If its premise is named, then the conjunct is given a conjunct identifier. Otherwise, 
   the first (second) branch of the dependency tree of its premise become the dependencies 
   of its conclusion.
 
 \item \texttt{GEN} and \texttt{SPEC}:
   The tags are unchanged by the application of those rules as they do not change the 
   structure of conjunctions. Although we have to be careful when using \texttt{SPEC} on 
   named theorems as it may create unwanted conjunctions. These virtual conjunctions are 
   not harmful as the right level of splitting is restored during the next phase.

\begin{example} (Creation of a virtual conjunction from a named theorem)
\begin{alltt}
\(\forall\) x.x \(\vdash\) \(\forall\) x.x 
                                   SPEC [A \(\wedge\) B]
\(\forall\) x.x \(\vdash\) A \(\wedge\) B 
                                   CONJUNCT1
\(\forall\) x.x \(\vdash\) A
\end{alltt}
\end{example}

 \item \texttt{SUBST}: Its premises consist of a theorem, a list of substitution theorems 
 of the form $(A=B)$ and a template that tells where each substitution should be applied. 
 When \texttt{SUBST} preserves
 the structure of conjuncts, the set of all identifiers in the passed dependencies of 
 the substitution theorems is distributed over each leaf of the tree given by the passed 
 dependencies of the substituted theorems. When it is not the case the dependency should be 
 flattened. Since the substitution of sub-terms below the top formula level does not affect 
 the structure of conjunctions, it is sufficient (although not necessary) to check that no 
 variables in the template is a predicate (is a boolean or returns a boolean).


\end{itemize}  

The heuristics presented above try to preserve the dependencies associated with single conjuncts whenever
possible. It is of course possible to find more advanced heuristics, that would give more precise human-proof
dependencies. However, performing more advanced operations (even pattern matching) may slow down the proof
system too much; so we decided to restrict to the above heuristics.

  Before exporting the theorems, we split them by recursively distributing 
  quantifiers and splitting conjunctions. This gives rise to conflicting degree of  
  splitting, as for instance, a theorem with many conjunctions may have been used as a 
  whole 
  during a proof. Given a theorem and its dependency tree, each of its conjunctions is 
  given the set of identifiers of its closest parent in this tree. Then, each of these 
  identifiers is also split maximally. In case of a virtual conjunction (see the \texttt{SPEC} 
  rule above), the corresponding node does not exist in the theorem tree, so we take the 
  conjunct corresponding to its closest parent. Finally, for each conjunct, we obtain a 
  set of dependencies by taking the union of the split identifiers.

\begin{example} (Recovering dependencies from the named theorem Th1)
\begin{alltt}
Th0 (named theorem): A \(\wedge\) B
Th1 (named theorem): C \(\wedge\) (D \(\wedge\) E)
     with dependency tree Tree([Th0],[Th0_c1])

  Recovering dependencies of each conjunct
Th1_c0: Th0 
Th1_c1: Th0_c1  
Th1_c2: Th0_c1
  Splitting the dependencies
Th1_c0: Th0_c1 Th0_c2
Th1_c1: Th0_c1 
Th1_c2: Th0_c1
\end{alltt}
\end{example}

\section{Evaluation}\label{sec:setting}

In this section we describe the setting used in the experiments: the ATPs,
the transformation from HOL to the formats of the ATPs, the dependencies
accessible in the different experiments, and the features used for machine learning.

\subsection{ATPs and problem transformation}

\holyhammer supports the translation to the formats of various TPTP ATPs:
FOF, TFF1, THF0, and two experimental TPTP extensions. In this paper
we restrict ourselves to the first order monomorphic logic, as these
ATPs have been the most powerful so far and integrating them in \holfour
already poses an interesting challenge. The transformation
that \holyhammer uses is heavily influenced by previous work by
Paulson~\cite{PaulsonS07} and Harrison~\cite{meson}. It is
described in detail in~\cite{holyhammer}, here we
remind only the crucial points. Abstractions are removed by $\beta$-reduction
followed by $\lambda$-lifting, predicates as arguments are removed by introducing
existentially quantified variables and the apply functor is used to reduce all
applications to first-order. By default \holyhammer uses the tagged polymorphic
encoding~\cite{BlanchetteBPS13}: a special tag taking two arguments is introduced,
and applied to all variable instances and certain applications. The first argument
is the first-order flattened representation of the type, with variables functioning
as type variables and the second argument is the value itself.

\begin{table}
  \centering
  \begin{tabular}{lcc}\toprule
Prover & Version & Premises \\ \midrule
\vampire & 2.6 & 96 \\
\eprover & 1.8 & 128 \\
\zthree  & 4.32 & 32 \\
\textsf{CVC4}      & 1.3 & 128 \\
\textsf{Spass}    & 3.5 & 32 \\
\textsf{IProver}  & 1.0 & 128 \\
\metis            & 2.3 & 32 \\ \bottomrule
  \end{tabular}
  \caption{\label{tab:provers}ATP provers, their versions and arguments}
\end{table}

The initially used provers, their versions and default numbers of premises are
presented in Table~\ref{tab:provers}.
The \hollight experiments~\cite{holyhammer} showed, that different provers
perform best with different given numbers of premises. This is particularly
visible for the ATP provers that already include the relevance filter SInE~\cite{sine},
therefore we preselect a number of predictions used with each prover. Similarly,
the strategies that the ATP provers implement are often tailored for best
performance on the TPTP library, for the annual CASC competition~\cite{Sutcliffe14}.
For ITP originating problems, especially for \eprover different strategies are
often better, so we run it under the alternate scheduler \texttt{Epar}~\cite{DBLP:journals/corr/abs-1301-2683}.

\subsection{Accessible facts}

As \holyhammer has initially been designed for \hollight, it treats
accessible facts in the same way as the accessibility relation defined
there: any fact that is present in a theory loaded chronologically
before the current one is available. In \holfour there are explicit
theory dependencies, and as such a different accessibility relation
is more natural. The facts present in the same theory before the current
one, and all the facts in the theories that the current one depends on
(possibly in a transitive way) are accessible. In this subsection
we discuss the four different accessible sets of lemmas, which we will
use to test the performance of \holyhammer on.

\paragraph{Exact dependencies (reproving)} They are the closest named ancestors of a theorem in the proof tree. It tests how much \holyhammer could reprove
if it had perfect predictions. In this settings no relevance filtering is done, as the 
number of dependencies is small.

\paragraph{Transitive dependencies} They are all the named ancestors of a theorem in the proof tree. It simulates proving a theorem in a perfect environment, where
all recorded theorems are a necessary step to prove the conjecture. This
corresponds to a proof assistant library that has been refactored into
little theories~\cite{Farmer92littletheories}.

\paragraph{Loaded theorems} All theorems present in the loaded theories are provided
together with all the theorems previously built in the current theory.
This is the setting used when proving theorems in \holfour, 
so it is the one we use in our interactive version presented and evaluated in Section \ref{sec:interactive}.

\paragraph{Linear order} For this experiment, we additionally recorded the order 
in which the \holfour theories were built,
so that we could order all the theorems of the standard library in a similar
way as \hollight theorems are ordered. All previously derived theorems are provided.

\subsection{Features}

Machine learning algorithms typically use features to define the
similarity of objects. In the large theory automated reasoning setting
features need to be assigned to each theorem, based on the
syntactic and semantic properties of the  statement of the theorem and
its attributes.

\holyhammer represents features by strings and characterizes theorems
using lists of strings. Features originate from the names of the
type constructors, type variables, names of constants and printed subterms
present in the conclusion. An important notion is the normalization of
the features: for subterms, their variables and type variables need to
be normalized. Various scenarios for this can be considered:

\begin{itemize}
\item All variables are replaced by one common variable.
\item Variables are replaced by their de Bruijn index numbers~\cite{US+08}.
\item Variables are replaced by their (variable-normalized) types~\cite{holyhammer}.
\end{itemize}

The union of the features coming from the three above normalizations has
been the most successful in the \hollight experiments, and it is used here as
well.

\subsection{Predictors}
In all our experiments we have used the modified k-NN algorithm~\cite{ckju-pxtp13}. This
algorithm produces the most precise results in the \holyhammer experiments for \hollight~\cite{holyhammer}.
Given a fixed number ($k$), the k-nearest neighbours learning algorithm finds $k$ premises
that are closest to the conjecture, and uses their weighted dependencies to find the predicted
relevance of all available facts. All the facts and the conjecture are interpreted as
vectors in the $n$-dimensional feature space, where $n$ is the number of all features.
The distance between a fact and the conjecture is computed using the Euclidean distance.
In order to find the neighbours of the conjecture efficiently, we store an association
list mapping features to theorems that have those features. This allows skipping the theorems
that have no features in common with the conjecture completely.

Having found the neighbours, the relevance of each available fact is computed by summing
the weights of the neighbours that use the fact as a dependency, counting each neighbour also as its own dependency

\section{Experiments}\label{sec:experiments}  
In this section, we present the results of several experiments and discuss the quality
of the advice system based on these results. The hardware used during the 
reproving and accessibility experiments is a 48-core server (AMD Opteron 6174 2.2 GHz.
CPUs, 320 GB RAM, and 0.5 MB L2 cache per CPU).
In these experiments, each ATPs is run on a single core for each problem with a time limit 
of 30 seconds. 
 The reconstruction and interactive experiments were run on a laptop with a Intel Core 
 processor (i5-3230M 4 x 2.60GHz with 3.6 GB RAM).

\subsection{Reproving}

We first try to reprove all the 9434 theorems in the \holfour libraries with the dependencies extracted from the proofs. This number is lower than the number of exported theorems because definitions are discarded. Table~\ref{Reproving_unsplit} presents the success rates for reproving using the dependencies recorded without
splitting. In this experiment we also compare many provers and their versions. For
\eprover~\cite{Schulz:LPAR-2013}, we also compare its different scheduling strategies \cite{DBLP:journals/corr/abs-1301-2683}. The results are used
to choose the best versions or strategies for the selected few provers. Apart from
the success rates, the unique number of problems is presented (proofs found by this ATP only), and \textsf{CVC4}~\cite{Barrett:2011:CVC:2032305.2032319} seems
to perform best in this respect. The translation used by default by \holyhammer is an
incomplete one (it gives significantly better results than complete ones), so some
of the problems are counter-satisfiable. 

\begin{table}[b!]
\centering
\begin{tabular}{@{}ccccc@{}}
\toprule
  Prover & Version & Theorem(\%)  & Unique & CounterSat \\
\midrule
  	\eprover & \texttt{Epar} 3 & 44.45  &	3 & 0\\
  	\eprover & \texttt{Epar} 1 & 44.15  & 	9 &	0\\
  	\eprover & \texttt{Epar} 2 & 43.95  & 	9 & 0\\	
  	\eprover & \texttt{Epar} 0 & 43.52  &	2 & 0\\	
  	\textsf{CVC4} & 1.3 & 42.71 &   44 & 0\\	
  	\zthree	& 4.32 & 41.96  & 8 &	5\\	
  	\zthree	& 4.40 & 41.65  & 1 &   6\\	
  	\eprover & 1.8 & 41.37  &   14 & 0\\
  	\vampire & 2.6 & 41.10  &   14 & 0\\
  	\vampire & 1.8 & 38.34  &	6 & 0\\
  	\zthree	& 4.40q & 35.19  &   11 & 5 \\
  	\vampire & 3.0 & 34.82  &    0 & 0\\
  	\textsf{Spass} & 3.5 & 31.67  &	0 & 0\\
  	\metis & 2.3 & 29.98 	& 0   & 0\\	
  	\textsf{IProver} & 1.0	& 25.52 	& 2	  & 35\\
    \midrule
    total   & & 50.96 & & 38\\\bottomrule
\end{tabular}
\caption{Reproving experiment on the 9434 unsplit theorems of the standard libary}
\label{Reproving_unsplit}
\end{table}

From this point on, experiments will be performed only with the best versions of three
provers: \eprover, \vampire~\cite{KovacsCAV13}, and \zthree~\cite{DeMoura:2008:ZES:1792734.1792766}. They have a high success rate
combined with an easy way of retrieving the unsatisfiable core. The same ones have been
used in the \holyhammer experiments for \hollight.

In Table~\ref{Reproving_split}, we try to reprove conjuncts of these theorems with the 
different recording methods described in Section~\ref{sec:record}. First, we notice that 
only \zthree benefits from the tracking of more accurate 
dependencies. More, removing the unnecessary conjuncts worsen the results of \eprover and 
\vampire. One reason is that \eprover and \vampire do well with large number of lemmas and 
although a conjunct was not used in the original proof it may well be useful to these 
provers.Suprisingly, the percentage of reproved facts did not increase compared to 
Table~\ref{Reproving_unsplit}, as this was the case for \hollight experiments. By looking 
closely at the data, we notice the presence of the \texttt{quantHeuristics} theory, where 
85 theorems are divided into 1538 conjuncts. As the percentage of reproving in this theory 
is lower than the average (16\%), the overall percentage gets smaller given the increased 
weight of this theory. Therefore, we have removed the quantHeuristic theory in the Basic* 
and Optimized* experiments for a fairer comparison with the previous experiments. 
Finally, if we compare the Optimized experiment with the similar \hollight reproving 
experiment on 14185 \textsf{Flyspeck} problems~\cite{holyhammer}, we notice that we can 
reprove three percent more theorems in \holfour. This is mostly due to a 10 percent 
increase in the performance of \zthree on \holfour problems.

\begin{table}[t!]
\centering
\begin{tabular}{@{}ccccc@{}}
\toprule
\phantom{abc} & Basic & Optimized & Basic* & Optimized* \\
\midrule
    \eprover & 42.43 & 42.41 & 46.23 & 45.91\\
    \vampire & 39.79 & 39.32 & 43.24 & 42.41\\ 
    \zthree  & 39.59 & 40.63 & 43.78 & 44.18\\
    \midrule
    total & 46.74 & 46.76 & 50.97 & 50.55\\\bottomrule
\end{tabular}
\caption{Success rates of reproving (\%) on the 13910 conjuncts of the standard library with different dependency tracking mechanism.}
\label{Reproving_split}
  \end{table}
  
  In Table~\ref{Reproving_theories} we have compared the success rates of reproving in different theories, as this
  may represent a relative difficulty of each theory and also the relative performance of 
  each 
  prover. We observe that \zthree performs best on the theories \texttt{measure} and 
  \texttt{probability}, \texttt{list} and \texttt{finite\_map}, whereas \eprover and 
  \vampire have a higher success rate on the theories \texttt{arithmetic}, \texttt{real}, 
  \texttt{complex} and \texttt{sort}. Overall, the high success rate in the 
  \texttt{arithmetic} and 
  \texttt{real} theories confirms that \holyhammer can already tackle this type of 
  theorems. 
  Nonetheless, it would still benefit from integrating more SMT-solvers' functionalities 
  on advanced theories based on \texttt{real} and \texttt{arithmetic}.  

\begin{table}[b!]
\centering

\begin{tabular}{@{}ccccc@{}}
\toprule
\phantom{ab} & \scriptsize{arith} & \scriptsize{real} & \scriptsize{compl} 
& \scriptsize{meas} \\
\midrule
\eprover &  61.29 & 72.97  & 91.22 & 27.01  \\
\vampire &  59.74 & 69.57  & 77.19 & 20.85   \\
\zthree &  51.42  & 64.46  & 86.84 & 31.27  \\\midrule
total &  63.63  & 75.31  & 92.10  & 32.70   \\
\bottomrule
\end{tabular}

\begin{tabular}{@{}ccccc@{}}
\toprule
\phantom{abc}  & \scriptsize{proba} & \scriptsize{list} & \scriptsize{sort} & \scriptsize{f\_map} \\
\midrule
\eprover & 42.16 & 23.56 & 34.54 &  33.07\\
\vampire & 37.34 & 21.96 & 32.72 & 27.16\\
\zthree & 54.21 & 25.62 & 25.45 &  43.70\\
\midrule
total & 55.42 & 26.77 & 40.00 & 45.27 \\
\bottomrule
\end{tabular}

\caption{Percentage (\%) of reproved theorems in the theories \texttt{arithmetic}, \texttt{real}, \texttt{complex}, \texttt{measure},  
\texttt{probability}, \texttt{list}, \texttt{sorting} and \texttt{finte\_map}. }
\label{Reproving_theories}
\end{table}

  \subsection{With different accessible sets}
    
  In Table~\ref{accessible_sets} we compare the quality of the predictions in 
  different proving environments. 
  We recall that only the transitive dependencies, loaded theories and 
  linear order settings are using predictions and that the number of these predictions is 
  adapted to 
  the ability of each provers. The exact dependencies setting (reproving), is copied from Table~\ref{Reproving_split}
  for easier comparison.

 \begin{table}[htb]
 \centering
 \begin{tabular}{@{}ccccc@{}}
 \toprule
 \phantom{abc} & ED & TD & LT & LO\\
 \midrule
   \eprover & 42.41 & 33.10 & 43.58 & 43.64\\
   \vampire & 39.32  & 29.56  & 38.46 & 38.54\\
   \zthree & 40.63  & 24.66 & 31.22 & 31.20\\
 \midrule
 total &  46.76 & 37.54  & 50.54 & 50.68 \\
 \bottomrule
 \end{tabular}
 \caption{Percentage (\%) of proofs found using different accessible sets: exact 
 dependencies (ED), transitive dependencies (TD), loaded theories (LT), and linear 
 order (LO)}
 \label{accessible_sets}
 \end{table} 
  
  We first notice the lower success rate in the transitive dependencies setting.
  There may be two justifications. First, the transitive 
  dependencies provide a poor training set for the predictors; the set of samples is 
  quite small and the available lemmas are all related to the conjecture. Second, it is very 
  unlikely that a lemma in this set will be better than a lemma in the exact dependencies, so we cannot 
  hope to perform better than in the reproving experiment.
  
  We now focus on the loaded theories and linear order settings, which are the two scenarios
  that correspond to the regular usage of a ``hammer'' system in a development: given all the
  previously known facts try to prove the conjecture. The results are surprisingly better
  than in the reproving experiment. First, this indicates 
  that the training data coming from a larger sample is better. Second, this shows that the \holfour 
  library is dense and that closer dependencies than the exact one may be found by the 
  predictors. It is quite common that large-theory automated reasoning techniques find
  alternate proofs.
  Third, if we look at each ATP separately, we see a one percent increase for \eprover, a 
  one percent decrease for \vampire, and 9 percent decrease for \zthree. This correlates
  with the number of selected premises. Indeed, it is easy to see that if a prover
  performs well with a large number of selected premises, it has more chance to find the
  relevant lemmas.
  Finally, we see that each of the provers enhanced the results by solving different problems.

  We can summarize the results by inferring that predictors combined with ATPs are most 
  effective in large and dense developments. 
  
  The linear order experiments was also designed to make a valid comparison with a similar  
  experiment where 39\% of \textsf{Flyspeck} theorems were proved by combining 14 methods This number was later raised to 47\% by improving the machine learning algorithm. Comparatively, the current 3 methods can prove 50\% of the \holfour theorems. This may be
  since the machine learning methods have improved, since the ATPs are stronger now or
  even because the \textsf{Flyspeck} theories contain a more linear (less dense) development
  than the \holfour libraries, which makes it harder for automated reasoning techniques.

  \subsection{Reconstruction}\label{sec:reconstruction}
Until now all the ATP proved theorems could only be used as oracles inside \holfour. This defeats 
the main aim of the ITP which is to guarantee the soundness of the 
proofs. The provers that we use in the experiments can return the unsatisfiable core: 
a small set of premises used during the proof. The HOL representation of these facts
 can be given to \metis in order to reprove the theorem with soundness guaranteed by its construction. We 
investigate reconstructing proofs found by \vampire on the loaded theories experiments 
(used in our interactive version of \holyhammer). We found that \metis could reprove, with 
a one second time limit, 95.6\% of these theorems. This result is encouraging for two reasons:
First, we have not shown the soundness of our transformations, and this shows that the found premises
indeed lead to a valid proof in HOL. Second, the high reconstruction rate suggest that the system
can be useful in practice.

\subsection{Case study}
Finally, we present two sets of lemmas found by \eprover advised on the loaded libraries. 
We discuss the difference with the lemmas used in the original proof.

The theorem \texttt{EULER\_FORMULE} states that any complex number can be represented as a 
combination of its norm and argument. In the human-written proof script ten
theorems are provided to a rewriting tactic. The user is mostly hindered by the fact that 
she could not use the commutativity of multiplication as the tactic would not terminate. 
Free of these constraints, the advice system returns only three lemmas: the commutativity 
of multiplication, the polar representation \texttt{COMPLEX\_TRIANGLE}, and the Euler's formula \texttt{EXP\_IMAGINARY}.

\begin{example}(In theory \texttt{complex})
\begin{alltt}
Original proof:
val EULER_FORMULE = store_thm("EULER_FORMULE",
  ``!z:complex. modu z * exp (i * arg z) = z``,
  REWRITE_TAC[complex_exp, i, complex_scalar_rmul, 
  RE, IM, REAL_MUL_LZERO, REAL_MUL_LID, EXP_0, 
  COMPLEX_SCALAR_LMUL_ONE, COMPLEX_TRIANGLE]);
              
Discovered lemmas:
COMPLEX_SCALAR_MUL_COMM COMPLEX_TRIANGLE 
EXP_IMAGINARY
\end{alltt}
\end{example}
 
The theorem \texttt{LCM\_LEAST} states that any number below the least common multiple is 
not a common multiple. This seems trivial but actually the least common multiple ($lcm$) 
of two natural numbers is defined as their product divided by their greatest common 
divisor. The user has proved the contraposition which requires two \metis 
calls. The discovered lemmas seem to indicate a similar proof, but it requires more 
lemmas, namely \texttt{FALSITY} and \texttt{IMP\_F\_EQ\_F} as the false constant is 
considered as any other constant in \holyhammer and uses the combination of 
\texttt{LCM\_COMM} and \texttt{NOT\_LT\_DIVIDES} instead 
of \texttt{DIVIDES\_LE}.

\nopagebreak{
\begin{example}(In theory \texttt{gcd}) 

\begin{alltt} 
Original proof:
val LCM_LEAST = store_thm("LCM_LEAST",
  ``0 < m \(\wedge\) 0 < n ==> !p. 0 < p \(\wedge\) p < lcm m n 
  ==> ~(divides m p) \(\vee\) ~(divides n p)``,
  REPEAT STRIP_TAC THEN SPOSE_NOT_THEN 
  STRIP_ASSUME_TAC THEN `divides (lcm m n) p` 
  by METIS_TAC [LCM_IS_LEAST_COMMON_MULTIPLE] 
  THEN `lcm m n <= p` by METIS_TAC [DIVIDES_LE] 
  THEN DECIDE_TAC);

Discovered lemmas: 
LCM_IS_LEAST_COMMON_MULTIPLE LCM_COMM 
NOT_LT_DIVIDES FALSITY IMP_F_EQ_F
\end{alltt}
\end{example} 
}

\subsection{Interactive version}\label{sec:interactive}
In our previous experiments, all the different steps (export, learning/predictions, 
translation, ATPs) were performed separately, and simultaneously for all the theorems. 
Here, we compose all this steps to produce one \holfour step, that given a conjecture 
proves it, usable in any \holfour development in an interactive advice loop. It proceeds
as follows: The conjecture is exported along with the 
currently loaded theories. Features for the theorems and the conjecture are computed, and
dependencies are used for learning and selecting the theorems relevant to the conjecture.
\holyhammer translates the problem to the formats of the ATPs and uses them to prove 
the resulting problems. If successful, the discovered unsatisfiable core, consisting of the 
\holfour theorems used in the ATP proof, is then read back to \holfour, returned as a proof 
advice, and replayed by \metis.

In the last experiment, we evaluate the time taken by each steps on two conjectures, which 
are not already proved in the \holfour libraries. The first tested goal $C_1$ is $gcd\ 
(gcd\ a\ a)\ (b + a) = (gcd\ b\ a)$, where $gcd\ n\ m$ is the greatest common divisor of 
$n$ and $m$. It can be automatically proved from three lemmas about $gcd$. The second goal 
is $C_2$ is $Im(i*i) = 0$, where $Im$ the imaginary part of a complex number. It can be 
automatically proved from 12 lemmas in the theories \texttt{real}, \texttt{transc} and \texttt{complex}.

In Table~\ref{tab:Timer}, the time taken by the export and import phase linearly depends on the number of theorems in the loaded libraries (given in parenthesis), as expected by the knowledge of our data and the complexity analysis of our code.

The time 
shown in the fourth column (``Predict'') includes the time to extract features, to learn 
from 
the dependencies and to find 96 relevant theorems. The time needed for machine learning
is relatively short. The time taken by \vampire shows that the 
second conjecture is harder. This is backed by the fact that we could not tell in advance 
what would be the necessary lemmas to prove this conjecture. The overall column presents 
the time between the 
interactive call and the display of advised lemmas. The low running times 
support the fact that our tool is fast enough for interactive use. 

\begin{table}[htb]
\centering
\begin{tabular}{@{}ccccccc@{}}
\toprule
\phantom{abc} & \small{Export} & \small{Import} & \small{Predict} & \small{\vampire} & 
\small{Total} \\
\midrule
$C_1$ (2224) & 0.38 & 0.20 & 0.29 & 0.01 & 0.97 \\
$C_2$ (4056) & 0.67 & 0.43 & 0.59 & 1.58 & 3.42 \\
\bottomrule
\end{tabular}
\caption{Time (in seconds) taken by each step of the advice loop}
\label{tab:Timer}
\end{table}

\section{Conclusion}\label{sec:concl}

In this paper we present an adaptation of the \holyhammer system for \holfour,
which allows for general purpose learning-assisted automated reasoning.
As \holyhammer uses machine learning for relevance filtering, we need to compute
the dependencies, define the accessibility relation for theorems and
adapt the feature extraction mechanism to \holfour. Further, as we export all
the proof assistant data (types, constants, named theorems) to a common format,
we define the namespaces to cover both \hollight and \holfour.

We have evaluated the resulting system on the \holfour standard library
toplevel goals: for about 50\% of them a sufficient set of dependencies
can be found automatically. We compare the success rates depending on the accessibility
relation and on the treatment of theorems whose statements are conjunctions.
We provide a \holfour command that translates
the current goal, runs premise selection and the ATP, and if a proof has
been found, it returns a \metis call needed to solve the goal.
The resulting system is available at \url{https://github.com/barakeel/HOL}.\marginpar

\subsection{Future Work}
The libraries of \hollight and \holfour are currently processed completely
independently. We have however made sure that all data is exported in the
same format, so that same concepts and theorems about them can be discovered
automatically~\cite{tgck-cicm14}. By combining the data, one might get goals
in one system solved with the help of theorems from the other, which can then
be turned into lemmas in the new system. A first challenge might be to
define a combined accessibility relation in order to evaluate such a combined
proof assistant library.

The format that we use for the interchange of \holfour and \hollight data is
heavily influenced by the TPTP formats for monomorphic higher-order logic~\cite{tptpthf0}
and polymorphic first-order logic~\cite{jasmincade2013}. It is however slightly
different from that used by Sledgehammer's \texttt{fullthf}. By completely
standardizing the format, it would be possible to interchange problems between
\sledgehammer and \holyhammer.

In \holfour, theorems include the information about the theory they originate
from and other attributes. It would be interesting to evaluate the impact
of such additional attributes used as features for machine learning on
the success rate of the proofs. Finally, most \holyhammer users call its web
interface~\cite{ckju-msc-hh-accepted}, rather than locally install the
necessary prover modifications, proof translation and the ATP provers. It would
be natural to extend the web interface to support \holfour.

\acks

We would like to thank Josef Urban and Michael F\"arber for their comments on the previous version
of this paper.
This work has been supported by the Austrian Science Fund (FWF): P26201.

\bibliographystyle{abbrvnat}
\bibliography{biblio}

\end{document}